\documentclass{article}

\usepackage{microtype}
\usepackage{graphicx}
\usepackage{subfigure}
\usepackage{booktabs} 
\usepackage{amsmath, dsfont, amsfonts}

\usepackage{hyperref}

\usepackage[accepted]{icml2018}

\icmltitlerunning{Structured Variational Learning of Bayesian Neural Networks with Horseshoe Priors}

\begin{document}

\twocolumn[
\icmltitle{Structured Variational Learning of Bayesian Neural Networks with Horseshoe Priors}



\begin{icmlauthorlist}
\icmlauthor{Soumya Ghosh}{to,ed}
\icmlauthor{Jiayu Yao}{goo}
\icmlauthor{Finale Doshi-Velez}{goo}
\end{icmlauthorlist}

\icmlaffiliation{to}{IBM research, Cambridge, MA, USA}
\icmlaffiliation{goo}{Harvard University, Cambridge, MA, USA}
\icmlaffiliation{ed}{MIT-IBM Watson AI Lab}

\icmlcorrespondingauthor{Soumya Ghosh}{ghoshso@us.ibm.com}

\icmlkeywords{Machine Learning, ICML}

\vskip 0.3in
]


\printAffiliationsAndNotice{}  

\begin{abstract}
  Bayesian Neural Networks (BNNs) have recently received increasing
  attention for their ability to provide well-calibrated posterior
  uncertainties.  However, model selection---even choosing the number
  of nodes---remains an open question.  Recent work has proposed the
  use of a horseshoe prior over node pre-activations of a Bayesian
  neural network, which effectively turns off nodes that do not help
  explain the data.  In this work, we propose several modeling and
  inference advances that consistently improve the compactness of the
  model learned while maintaining predictive performance, especially
  in smaller-sample settings including reinforcement learning.
\end{abstract}

\newcommand{\Xmat}{\text{X}}
\newcommand{\Ymat}{\text{Y}}
\newcommand{\W}[1]{\mathcal{W}_{#1}}
\newcommand{\s}{\mathcal{S}}
\newcommand{\ind}[1]{\mathbf{1}[#1]}
\newcommand{\wvar}[1]{\tau_#1^{-1}}
\newcommand{\data}{\mathcal{D}}
\newcommand{\eye}{\mathbb{I}}
\newcommand{\T}{\mathcal{\theta}}
\newcommand{\Tau}{\mathcal{T}}
\newcommand{\blayer}{b_g}
\newcommand{\bnode}{b_0}
\newcommand{\invgamma}{\text{Inv-Gamma}}
\newcommand{\normal}{\mathcal{N}}
\newcommand{\halfcauchy}{C^{+}}
\newcommand{\taunode}{\tau_{kl}}
\newcommand{\tildetaunode}{{\tilde{\tau}}_{kl}}
\newcommand{\taulayer}{\upsilon_{l}}
\newcommand{\lambdanode}{\lambda_{kl}}
\newcommand{\lambdalayer}{\vartheta_{l}}
\newcommand{\lambdakappa}{\rho_{\kappa}}
\newcommand{\unode}{u_{kl}}
\newcommand{\wnode}{w_{kl}}
\newcommand{\betanode}{\beta_{kl}}
\newcommand{\half}{\frac{1}{2}}
\newcommand{\elbo}{\mathcal{L}}
\newcommand{\gradelbo}{\nabla_\phi\hat{\mathcal{L}}(\phi)}
\newcommand{\E}[1]{\mathbb{E}[#1]}
\newcommand{\Ewrt}[2]{\mathbb{E}_{#1}[#2]}
\newcommand{\ent}[1]{\mathbb{H}[#1]}
\newcommand{\real}[1]{\mathbb{R}^{#1}}
\newcommand{\bkappa}{b_\kappa}
\newcommand{\betamat}{\beta_{l}}
\newcommand{\mvn}{\mathcal{M}\mathcal{N}}
\newcommand{\nusample}{\nu_l^{(s)}}
\newcommand{\cbetamat}{M_{{\betamat}\mid\nu_l}}
\newcommand{\cU}{U_{{\betamat}\mid\nu_l}}
\newcommand{\vecB}{\vec{B}}
\newcommand{\vecM}{\vec{M}}
\newcommand{\cbetaj}{{\beta_j}\mid{\nu_j}}
\section {Introduction}
Bayesian Neural Networks (BNNs) are increasingly the de-facto approach
for modeling stochastic functions.  By treating the weights in a
neural network as random variables, and performing posterior inference
on these weights, BNNs can avoid overfitting in the regime of small
data, provide well-calibrated posterior uncertainty estimates, and
model a large class of stochastic functions with heteroskedastic and
multi-modal noise.  These properties have resulted in BNNs being
adopted in applications ranging from active learning~\cite{MHLobato15,
  YGal16Active} and reinforcement
learning~\cite{CBlundell15,depeweg2016learning}.
\begin{figure}[t]
\centering 
\includegraphics[width=\columnwidth]{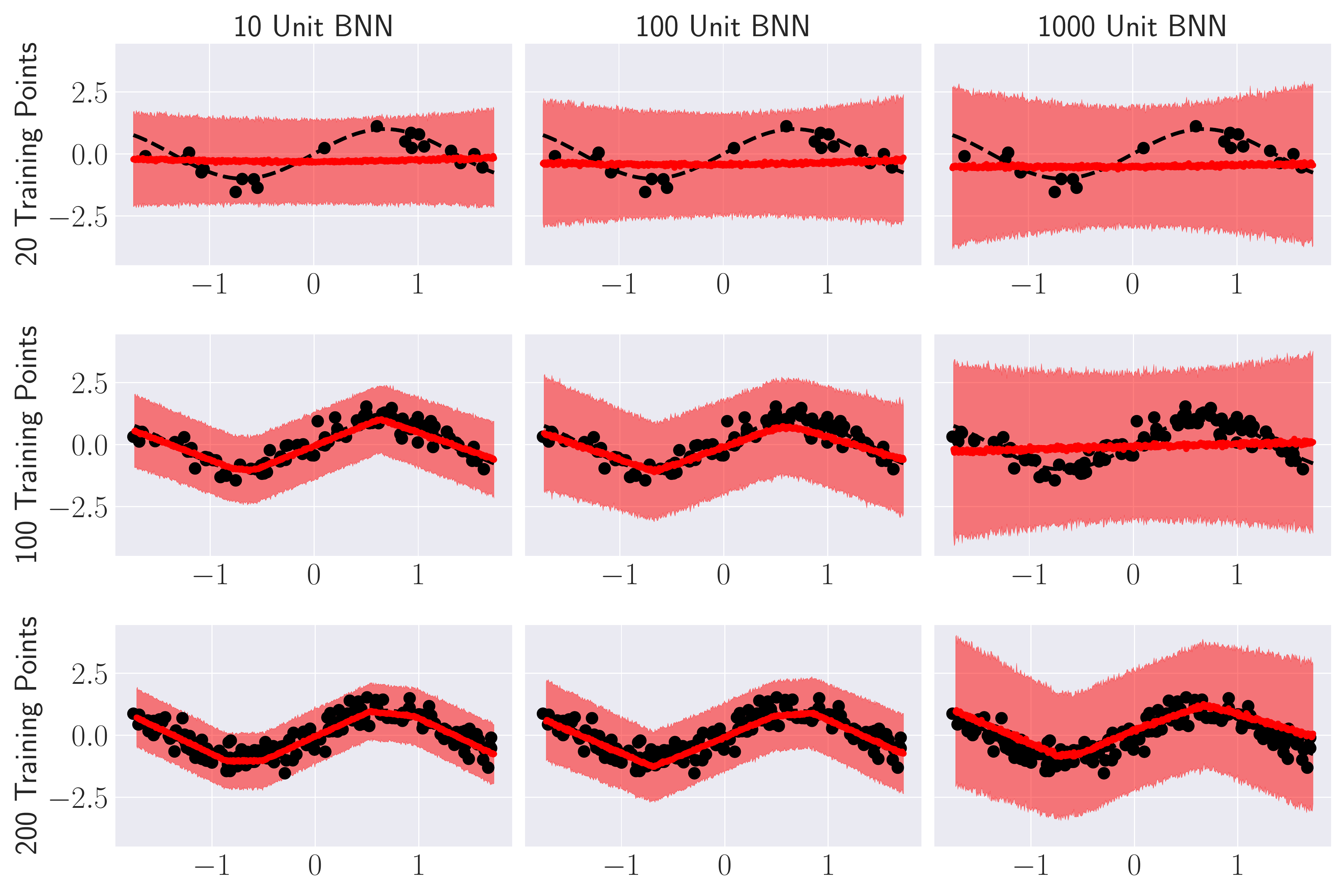}
\vskip -0.2in
\caption{\small{Predictive distributions from a single layer BNN with $\normal(0, 1)$ priors over weights, containing 10, 100, and 1000 units, trained on noisy samples (in black) from a smooth 1 dimensional function shown in black. With fixed data increasing BNN capacity leads to over-inflated uncertainty.}}	
\label{fig:uncertainty}
\end{figure}

While there have been many recent advances in training
BNNs~\cite{MHLobato15, CBlundell15, DRezende14, CLouizos16,
  MHLobato16}, model-selection in BNNs has received relatively less
attention.  Unfortunately, the consequences for a poor choice of
architecture are severe: too few nodes, and the BNN will not be
flexible enough to model the function of interest; too many nodes, and
the BNN predictions will have large variance.  We note that these
Bayesian model selection concerns are subtlely different from
overfitting and underfitting concerns that arise from maximum
likelihood training: here, more expressive models (e.g. those with
more nodes) require more data to concentrate the posterior.  When
there is insufficent data, the posterior uncertainty over the BNN
weights will remain large, resulting in large variances in the BNN's
predictions.  We illustrate this issue in
Figure~\ref{fig:uncertainty}, where we see a BNN trained with too many
parameters has higher variance around its predictions than one with
fewer.  Thus, the core concern of Bayesian model selection is to
identify a model class expressive enough that it can explain the
observed data set, but not so expressive that it can explain
everything \cite{rasmussen2001occam, murray2005note}.

Model selection in BNNs is challenging because the number of nodes in
a layer is a discrete quantity.  Recently, \cite{SGhosh17,
  louizos2017bayesian} independently proposed performing model
selection in Bayesian neural networks by placing Horseshoe
priors~\cite{CCarvalho09} over the weights incident to
each node in the network.  This prior can be interpreted as a
continuous relaxation of a spike-and-slab approach that would assign a
discrete on-off variable to each node, allowing for
computationally-efficient optimization via variational inference.

In this work, we expand upon this idea with several innovations and careful
experiments.  Via a combination of using regularized horseshoe priors
for the node-specific weights and variational
approximations that retain critical posterior structure, we both improve upon the statistical properties of the earlier works and provide improved generalization, especially for smaller data sets and in sample-limited settings such as reinforcement learning. We also present a new thresholding rule for pruning away nodes. Unlike previous work our rule does not require computing a point summary of the inferred posteriors. We compare the various model and inference combinations on a diverse set of regression and reinforcement learning tasks. We find that the proposed innovations consistently improve upon the compactness of the models learned without sacrificing predictive performance. 
\section {Bayesian Neural Networks}
A Bayesian neural network endows the parameters $\W{}$ of a neural
network with distributions $\W{} \sim p(\W{})$. When combined with
inference algorithms that infer posterior distributions over weights,
they are able to capture posterior as well as predictive
uncertainties.  For the following, consider a fully connected deep
neural network with $L-1$ hidden layers, parameterized by a set
of weight matrices $\W{} = \{W_l\}_{1}^{L}$, where $W_l$ is of size
$\real{K_{l-1} + 1\times K_l}$, and $K_l$ is the number of units in
layer $l$.  The network maps an input $x \in \real{D}$ to a response
$f(\W{}, x)$ by recursively applying the transformation
$h(W_l^T[z_l^T, 1]^T)$, where $z_l \in \real{K_l\times 1}$ is the
input into layer $l$, the initial input $z_0$ is $x$, and $h$ is a
point-wise non-linearity such as the rectified-linear function, $h(a)
= \text{max}(0, a)$.

Given $N$ observation response pairs $\data
= \{x_n, y_n\}_{n=1}^{N}$ and $p(\W{})$, we are interested in the posterior distribution
$p(\W{}\mid \data) \propto \prod_{n=1}^N p(y_n\mid f(\W{}, x_n))p(\W{}),
$
and in using it for predicting responses to unseen
data $x_*$,
$p(y_* \mid x_*) = \int p(y_*\mid f(\W{},x_*))p(\W{}\mid\data) d\W{}$.	
The prior $p(\W{})$ allows one to encode problem-specific beliefs as well as general properties about weights.
\section{Bayesian Neural Networks with Regularized Horseshoe Priors}
Let $\wnode \in \real{K_{l-1}+1 \times 1}$ denote the set of all weights incident into unit $k$ of hidden layer $l$.  \citet{SGhosh17,
  louizos2017bayesian} introduce a prior such that each unit's weight vector $\wnode$ is conditionally independent and follow a group Horseshoe prior~\cite{CCarvalho09},
\begin{eqnarray}
  \wnode \mid \taunode, \taulayer \sim \mathcal{N}(0, (\taunode^2\taulayer^2)\eye), \nonumber \\ \taunode \sim \halfcauchy(0, \bnode), \quad \taulayer \sim \halfcauchy(0, \blayer).
\label{eq:HS}
\end{eqnarray}
Here, $\eye$ is an identity matrix, $a \sim \halfcauchy(0, b)$ is the Half-Cauchy distribution with density $p(a|b) = 2/\pi b(1 + (a^2/b^2))$ for $a>0$, $\taunode$ is a unit specific scale parameter, while the scale parameter $\taulayer$ is shared across the layer.  This horseshoe prior exhibits Cauchy-like flat, heavy tails while maintaining an infinitely tall spike at zero. As a result, it allows sufficiently large unit weight vectors $\wnode$ to escape un-shrunk---by having a large scale parameter---while providing severe shrinkage to small weights. By forcing all weights incident on a unit to share scale parameters, we are able to induce sparsity at the unit level, turning off units that are unnecessary for explaining the data well. Intuitively, the shared layer wide scale $\taulayer$ pulls all units in layer $l$ to zero, while the heavy tailed unit specific $\taunode$ scales allow some of the units to escape the shrinkage.

\paragraph{Regularized Horseshoe Priors}
While the horseshoe prior has some good properties, when the amount of training data is limited, units with essentially no shrinkage can produce large weights can adversely affect generalization performance of HS-BNNs,  with minor perturbations of the data leading to vastly different predictions.  To deal with this issue, here we consider the regularized horseshoe prior~\cite{Juho17}. Under this prior $\wnode$ is drawn from,
\begin{equation}
	\wnode \mid \taunode, \taulayer, c \sim \mathcal{N}(0, (\tildetaunode^2\taulayer^2)\eye),  \tildetaunode^2 = \frac{c^2\taunode^2}{c^2 + \taunode^2\taulayer^2}.
	\label{eq:reg-HS}
\end{equation}
Note that for the weight node vectors that are strongly shrunk to zero, we will have tiny $\taunode^2\taulayer^2$. When, $\taunode^2\taulayer^2 \ll c^2$,  $\tildetaunode^2 \rightarrow \taunode^2\taulayer^2$, recovering the original horseshoe prior. On the other hand, for the un-shrunk weights $\taunode^2\taulayer^2$ will be large, and when $\taunode^2\taulayer^2 \gg c^2$, $\tildetaunode^2 \rightarrow c^2$. Thus, these weights under the regularized Horseshoe prior follow $\wnode\sim\normal(0, c^2\eye)$ and $c$ acts as a weight decay hyper-parameter. We place a $\invgamma(c_a, c_b)$ prior on $c^2$. In the experimental section, we find that the regularized HS-BNN does indeed improve generalization over HS-BNN.  Below, we describe two essential parametrization considerations essential for using the regularized horseshoe in practice. 

\emph{Half-Cauchy re-parameterization for variational learning.}
Instead of directly parameterizing the Half-Cauchy random variables in Equations~\ref{eq:HS} and~\ref{eq:reg-HS}, we use a convenient auxiliary variable parameterization~\cite{MWand11} of the distribution, $a \sim \halfcauchy(0, b)	\Longleftrightarrow a^2 \mid \lambda \sim \invgamma(\frac{1}{2}, \frac{1}{\lambda}); \lambda \sim \invgamma(\frac{1}{2}, \frac{1}{b^2}),$
where $v \sim \invgamma(a ,b)$ is the Inverse Gamma distribution with density $p(v) \propto v^{-a-1}\text{exp}\{-b/v\}$ for $v>0$. This avoids the challenges posed by the direct approximation during variational learning --- standard exponential family variational approximations struggle to capture the thick Cauchy tails, while a Cauchy approximating family leads to high variance gradients.

Since the number of output units is fixed by the problem at hand, a sparsity inducing prior is not appropriate for the output layer. Instead, we place independent Gaussian priors, $w_{kL} \sim \mathcal{N}(0, \kappa^2\eye)$ with vague hyper-priors $\kappa \sim \halfcauchy(0, \bkappa=5)$ on the output layer weights.  The joint distribution of the regularized Horseshoe Bayesian neural network is then given by,
\begin{equation}
\small
\begin{split}
&p(\data, \theta) = p(c\mid c_a, c_b)r(\kappa, \lambdakappa \mid \bkappa)\prod_{k=1}^{K_L}\mathcal{N}(w_{kL} \mid 0, \kappa\eye) \\
&\prod_{l=1}^{L} r(\taulayer, \lambdalayer \mid \blayer) 
\prod_{k=1}^{K_l}r(\taunode, \lambdanode \mid \bnode)\normal(\wnode\mid 0, (\tildetaunode^2\taulayer^2)\eye) \\
&\prod_{n=1}^N p(y_n\mid f(\W{}, x_n)),	
\end{split}
\end{equation}
where $ p(y_n|f(\W{}, x_n))$ is the likelihood function and $r(a, \lambda | b) =\invgamma(a^2|\frac{1}{2}, \frac{1}{\lambda})\invgamma(\lambda|\frac{1}{2}, \frac{1}{b^2})$,
with $\T = \{\W{}, \mathcal{T}, \kappa, \lambdakappa, c\}$, $\mathcal{T} =\{ \{\taunode\}_{k=1,l=1}^{K,L}, \{\taulayer\}_{l=1}^L, \{\lambdanode\}_{k=1,l=1}^{K,L}, \{\lambdalayer\}_{l=1}^L\}$. 

\emph{Non-Centered Parameterization} 
The regularized horseshoe (and the horseshoe) prior both exhibit strong correlations between the  weights $\wnode$ and the scales $\taunode\taulayer$. While their favorable sparsity inducing properties stem from this coupling, it also gives rise to coupled posteriors that exhibit pathological funnel shaped geometries~\cite{MBetancourt15, JIngraham16} that are difficult to reliably sample or approximate.  

Adopting non-centered parameterizations~\cite{JIngraham16}, helps alleviate the issue. Consider a reformulation of Equation~\ref{eq:reg-HS}, 
\begin{equation}   
\beta_{kl} \sim \normal(0, \eye)	, \quad \wnode = \tildetaunode\taulayer\beta_{kl},
\label{eq:ncp}
\end{equation}
where the distribution on the scales are left unchanged. Since the scales and weights are sampled from independent prior distributions and are \emph{marginally} uncorrelated, such a parameterization is referred to as non-centered. The likelihood is now responsible for introducing the coupling between the two, when conditioning on observed data. Non-centered parameterizations are known to lead to simpler posterior geometries~\cite{MBetancourt15}. Empirically~\cite{SGhosh17} have shown that adopting a non-centered parameterization significantly improves the quality of the posterior approximation for BNNs with Horseshoe priors.  Thus, we also adopt non-centered parameterizations for the regularized Horseshoe BNNs.

\section{Structured Variational Learning of Regularized Horseshoe BNNs}
We approximate the intractable posterior $p(\T \mid \data)$ with a computationally convenient family. We exploit recently proposed stochastic extensions to scale to both large architectures and datasets, and use black-box variants to deal with non-conjugacy. 
We begin by selecting a tractable family of distributions $q(\T \mid \phi)$, with free variational parameters $\phi$. Learning involves optimizing $\phi$ such that the Kullback-Liebler divergence between the approximation and the true posterior, $\text{KL}(q(\theta\mid \phi) || p(\theta \mid \data))$ is minimized.  This is equivalent to maximizing the lower bound to the marginal likelihood (or evidence) $p(\data)$, $p(\data) \geq \elbo(\phi) = \Ewrt{q_\phi}{\text{ln } p (\data, \theta)} + \ent{q(\theta\mid\phi)}$.	The choice of the approximating family governs the quality of inference.  
\subsection{Variational Approximation Choices}
\label{sec:v_families}
The more flexible the approximating family the better it approximates the true posterior.  Below, we first describe a straight-forward fully-factored approximation and then a more sophisticated structured approximation that we demonstrate has better statistical properties.  

\paragraph{Fully Factorized Approximations} The simplest possibility is to use a fully factorized variational family, 
\begin{equation}
\small
\begin{split}
&q(\T \mid \phi) = \prod_{a\in \{c, \kappa, \lambdakappa \}}q(a\mid\phi_a)\prod_{i,j,l}q(\beta_{ij,l}\mid \phi_{\beta_{ij,l}}) \\
&\prod_{k,l} q(\taunode\mid\phi_{\taunode})q(\lambdanode\mid\phi_{\lambdanode}) \prod_l q(\taulayer\mid\phi_{\taulayer})q(\lambdalayer\mid\phi_{\lambdalayer}).
\end{split}
\label{eq:approx_family}
\end{equation}
Restricting the variational distribution for the non-centered weight $\beta_{ij,l}$ between units $i$ in layer $l-1$ and $j$ in layer $l$, $q(\beta_{ij,l}\mid \phi_{\beta_{ijl}})$ to the Gaussian family $\normal(\beta_{ij,l}\mid \mu_{ij,l},\sigma^2_{ij,l})$, and the non-negative scale parameters $\taunode^2$ and $\taulayer^2$ and the variance of the output layer weights to the log-Normal family, $q(\text{ln } \taunode^2\mid \phi_{\taunode})  = \normal(\mu_{\taunode}, \sigma^2_{\taunode})$, $q(\text{ln } \taulayer^2 \mid \phi_{\taulayer}) = \normal(\mu_{\taulayer}, \sigma^2_{\taulayer})$, and $q(\text{ln } \kappa^2 \mid \phi_{\kappa}) = \normal(\mu_{\kappa}, \sigma^2_{\kappa})$, allows for the development of straightforward inference algorithms~\cite{SGhosh17, louizos2017bayesian}. It is not necessary to impose distributional constraints on the variational approximations of the auxiliary variables $\lambdalayer$, $\lambdanode$, or $\lambdakappa$. Conditioned on the other variables the optimal variational family for these latent variables follow inverse Gamma distributions. We refer to this approximation as the \emph{factorized} approximation. 
%

\emph{Parameter-tied factorized approximation.}  The conditional variational distribution on $\wnode$ implied by Equations~\ref{eq:approx_family} and~\ref{eq:reparam} is $q(\wnode \mid \taunode, \taulayer) = \mathcal{N}(\wnode \mid \taunode\taulayer\mu_{kl}, (\taunode\taulayer)^2\Psi)$, where $\Psi$ is a diagonal matrix with elements populated by $\sigma^2_{ij,l}$ and $\mu_{kl}$ consists of the corresponding variational means $\mu_{ij, l}$. The distributions of weights incident into a unit are thus coupled through $\taunode\taulayer$ while all weights in a layer are coupled through the layer wise scale $\taulayer$. This view suggests that using a simpler approximating family $q(\beta_{ij,l}\mid \phi_{\beta_{ijl}}) = \normal(\beta_{ij,l}\mid \mu_{ij,l}, 1)$ results in an isotropic Gaussian approximation $q(\wnode \mid \taunode, \taulayer) = \normal(\wnode \mid \taunode\taulayer\mu_{kl}, (\taunode\taulayer)^2\eye)$. Crucially, the scale parameters $\taunode\taulayer$ still allow for pruning of units when the scales approach zero. Moreover, by tying the variances of the non-centered weights together this approximation effectively halves the number of variational parameters and speeds up training~\cite{SGhosh17}. We call this the \emph{tied-factorized} approximation.
\paragraph{Structured Variational Approximations}
Although computationally convenient, the factorized approximations fail to capture posterior correlations among the network weights, and more pertinently, between weights and scales. 

We take a step towards a more structured variational approximation by using a layer-wise matrix variate Gaussian variational distribution for the non-centered weights and retaining the form of all the other factors from Equation~\ref{eq:approx_family}. Let $\beta_l \in \real{K_{l-1} + 1\times K_{l}}$ denote the set of weights betweens layers $l-1$ and $l$, then under this variational approximation we have $q(\betamat \mid \phi_{\betamat}) = \mvn(\betamat \mid M_{\betamat}, U_{\betamat}, V_{\betamat})$, where $M_{\betamat} \in \real{K_{l-1} + 1\times K_l}$ is the mean, $V_{\betamat} \in \real{K_l \times K_l}$ and $U_{\betamat} \in \real{K_{l-1} + 1 \times K_{l-1} + 1}$ capture the covariances among the columns and rows of $\betamat$, thereby modeling dependencies among the variational approximation to the weights in a layer. \citet{CLouizos16} demonstrated that even when $U_{\betamat}$ and $V_{\betamat}$ are restricted to be diagonal, the matrix Gaussian approximation can lead to significant improvements over fully factorized approximations for vanilla BNNs.  We call this the \emph{semi-structured}\footnote{it captures correlations among weights but not between weights and scales} approximation.

The horseshoe prior exhibits strong correlations between weights and their scales, which encourages strong posterior coupling between $\betanode$ and $\taunode$. For effective shrinkage towards zero, it is important that the variational approximations are able to capture this strong dependence. To do so, let $ B_l=
\left[
   \begin{array}{c}
     \betamat \\
	\nu_{l}^T
    \end{array}
\right ]$, $\nu_l = [\nu_{1l}, \ldots, \nu_{{K_l}l}]^T$, and $\nu_{kl} = \text{ln}\taunode$. Now using the variational approximation $q(B_l\mid\phi_{B_l}) = \mvn(B_l\mid M_l, U_l, V_l)$, allows us to retain the coupling between weights incident into a unit and the corresponding unit specific scales, with appropriate parameterizations of $U_l$. In particular, we note that a diagonal $U_l$ fails to capture the necessary correlations, and defeats the purpose of using a matrix Gaussian variational family to model the posterior of $B_l$. To retain computational efficiency while capturing dependencies among the rows of $B_l$ we enforce a low-rank structure, $U_l =  \Psi_l + h_lh_l^T$, where $\Psi_l  \in \real{K_{l-1} + 2\times K_{l-1} + 2}$ is a diagonal matrix and $h_l \in \real{K_{l-1} + 2 \times 1}$ is a column vector. We retain a diagonal structure for $V_l \in \real{K_l \times K_l}$. We call this approximation the \emph{structured} approximation. In the experimental section, we find that this structured approximation, indeed leads to stronger shrinkage towards zero in the recovered solutions. When combined with a pruning rule, it significantly compresses networks with excess capacity.
%
\begin{table}[t]
\caption{Variational Approximation Families.}
\label{tab:summary}
\begin{center}
\begin{small}
\begin{sc}
\resizebox{1.0\columnwidth}{!}{%
\begin{tabular}{ll}
\toprule
Approximation & Description  \\
\midrule
\small{Factorized}    & $\displaystyle q(\nu_l\mid\phi_{\nu_l})q(\betamat\mid\phi_{\betamat}) =\prod_{i,j,l}\normal(\betanode\mid \mu_{ij,l}, \sigma^2_{ij,l})\prod_{k, l}q(\nu_{kl}\mid\phi_{\nu_{kl}}) $ \\
\small{Factorized (tied)} & $\displaystyle
q(\nu_l\mid\phi_{\nu_l})q(\betamat\mid\phi_{\betamat}) = \prod_{i,j,l}\normal(\betanode\mid\mu_{ij,l}, 1) \prod_{k, l}q(\nu_{kl}\mid\phi_{\nu_{kl}})$ \\
\small{Semi-structured}    & $\displaystyle q(\nu_l\mid\phi_{\nu_l})q(\betamat \mid \phi_{\betamat}) = \mvn(\betamat \mid M_{\betamat}, U_{\betamat}, V_{\betamat})\prod_{k, l}q(\nu_{kl}\mid\phi_{\nu_{kl}})$ \\
\small{Structured}    & $ \displaystyle q(\betamat, \nu_l\mid\phi_{B_l}) = \mvn(B_l\mid M_l, U_l, V_l)$          \\
\end{tabular}%
}
\end{sc}
\end{small}
\end{center}
\vskip -0.1in
\end{table}
Table~\ref{tab:summary} summarizes the variational approximations introduced in this section.
\subsection{Black Box Variational Inference} 
Irrespective of the variational family choice, the resulting evidence lower bound (ELBO),
\begin{equation}
\small
  \begin{split}
\small
	&\elbo(\phi) = \sum_n \E{\text{ln } p(y_n \mid f(\beta, \Tau, \kappa, x_n))} + 
	\\ &\E{\text{ln } p(\Tau, \beta, \kappa, \lambdakappa \mid \bnode, \blayer, \bkappa)} +  \ent{q(\theta\mid \phi)},
  \end{split}
\label{eq:elbo}
\end{equation}
is challenging to evaluate. Here we have used $\beta$ to denote the set of all non-centered weights in the network. The non-linearities introduced by the
neural network and the potential lack of conjugacy between the neural
network parameterized likelihoods and the Horseshoe priors render the
first expectation in Equation~\ref{eq:elbo} intractable. 

Recent progress in black box variational inference~\cite{DKingma14, DRezende14, RRanganath14, MTitsias14} subverts this difficulty. These techniques compute noisy unbiased estimates of the gradient $\gradelbo$, by approximating the offending expectations with unbiased Monte-Carlo estimates and relying on either score function estimators~\cite{RWilliams92, RRanganath14} or reparameterization gradients~\cite{DKingma14, DRezende14, MTitsias14} to differentiate through the sampling process. With the unbiased gradients in hand, stochastic gradient ascent can be used to optimize the ELBO. In practice, reparameterization gradients exhibit significantly lower variances than their score function counterparts and are typically favored for differentiable models. The reparameterization gradients rely on the existence of a parameterization that separates the source of randomness from the parameters with respect to which the gradients are sought. For our Gaussian variational approximations, the well known non-centered parameterization, $\zeta \sim \mathcal{N}(\mu, \sigma^2) \Leftrightarrow \epsilon \sim \mathcal{N}(0, 1), \zeta = \mu + \sigma \epsilon$,  allows us to compute Monte-Carlo gradients,
\begin{equation}
\small
\begin{split}
&\nabla_{\mu, \sigma} \mathbb{E}_{q_w}[g(w)] \Leftrightarrow \nabla_{\mu, \sigma} \mathbb{E}_{\mathcal{N}(\epsilon\mid0,1)}[g(\mu + \sigma \epsilon)] \\
&\approx \frac{1}{S} \sum_s \nabla_{\mu, \sigma} g(\mu + \sigma \epsilon^{(s)}),\end{split}
\label{eq:reparam}
\end{equation}
for any differentiable function $g$ and $\epsilon^{(s)} \sim \mathcal{N}(0,1)$. Furthermore, all practical implementations of variational Bayesian neural networks use a further re-parameterization to lower variance of the gradient estimator. They sample from the implied variational distribution over a layer's pre-activations instead of directly sampling the much higher dimensional weights~\cite{DKingma15}.

\textbf{Variational distribution on pre-activations}
The ``local'' re-parametrization is straightforward for all the approximations except the structured approximation.  For that, observe that $q(B_l\mid\phi_{B_l})$ factorizes as $q(\betamat \mid  \nu_l, \phi_{\betamat})q(\nu_l \mid \phi_{\nu_l})$. Moreover, conditioned on $\nu_l \sim q(\nu_l \mid \phi_{\nu_l})$,  $\betamat$ follows another matrix Gaussian distribution. The conditional variational distribution is $q(\betamat \mid  \nu_l, \phi_{\betamat}) = \mvn(\cbetamat, \cU, V)$. It then follows that $b = \betamat^Ta$ for an input $a \in \real{K_{l-1} + 1\times 1}$ into layer $l$, is distributed as, 
\begin{equation}
\small
\begin{split}
b \mid a, \nu_l, \phi_{\betamat} \sim \normal(b \mid \mu_b, \Sigma_b),
 \end{split}
 \label{eq:pre}
\end{equation}
with $\mu_b= \cbetamat^T a$, and $ \Sigma_b = (a^T\cU a)V$. Since, $a^T\cU a$ is scalar and $V$ is diagonal, $\Sigma$ is diagonal as well. 
For regularized HS-BNN, recall that the pre-activation of node $k$ in layer $l$, is $\unode = \tildetaunode\taulayer b$, and the corresponding variational posterior is,
\begin{equation}
\small
\begin{split}
q(\unode\mid \mu_{\unode}, \sigma^2_{\unode}) = \normal(\unode\mid \mu_{\unode}, \sigma^2_{\unode}), \\
\mu_{\unode} = \tildetaunode^{(s)}\taulayer^{(s)}\mu_{bk}; \quad \sigma^2_{\unode} = {\tildetaunode^{(s)^2}}{\taulayer^{(s)}}^2\Sigma_{b k,k},
\end{split} 
\label{eq:lrpm}
\end{equation}
where $\taunode^{(s)}$, $\taulayer^{(s)}, c^{(s)}$ are samples from the corresponding log-Normal posteriors and $\tildetaunode^{(s)}$ is constructed as ${c^{(s)}}^2{\taunode^{(s)}}^2/({c^{(s)}}^2 + {\taunode^{(s)}}^2{\taulayer^{(s)}}^2)$. 
\\
 
%
\textbf{Algorithm} We now have a simple prescription for optimizing Equation~\ref{eq:elbo}.  Recursively sampling the variational posterior of Equation~\ref{eq:lrpm} for each layer of the network, allows us to forward propagate information through the network. Using the reparameterizations (Equation~\ref{eq:reparam}), allows us to differentiate through the sampling process. We compute the necessary gradients through reverse mode automatic differentiation tools~\cite{DMaclaurin15}. With the gradients in hand, we optimize $\elbo(\phi)$  with respect to the variational weights $\phi_{B}$, per-unit scales $\phi_{\taunode}$, per-layer scales $ \phi_{\taulayer}$, and the variational scale for the output layer weights, $\phi_\kappa$ using Adam~\cite{DKingma2014adam}. Conditioned on these, the optimal variational posteriors of the auxiliary variables $\lambdalayer$, $\lambdanode$, and $\lambdakappa$ follow Inverse Gamma distributions. Fixed point updates that maximize $\elbo(\phi)$ with respect to $\phi_{\lambdalayer}, \phi_{\lambdanode}, \phi_{\lambdakappa}$, holding the other variational parameters fixed are available. It can be shown that,
$q(\lambdanode \mid \phi_{\lambdanode}) = \invgamma(\lambdanode \mid 1, \E{\frac{1}{\taunode}} + \frac{1}{\bnode^2})$.
The distributions of the other auxiliary variables are analogous. By alternating between gradient and fixed point updates to maximize the ELBO in a coordinate ascent fashion we learn all variational parameters jointly (see Algorithm 1 of the supplement). Further details are available in the supplement.
%

\textbf{Computational Considerations}
The primary computational bottleneck for the structured approximation arises in computing the pre-activations in equation~\ref{eq:pre}. While computing $\Sigma_b$ in the factorized approximation involves a single inner product, in the structured case it requires the computation of the quadratic form $a^TU_{\cbetamat} a$ and a point wise multiplication with the elements of $V_l$. Owing to the diagonal plus rank-one structure of $U_{\cbetamat}$, we only need two inner products, followed by a scalar squaring and addition to compute the quadratic form and $K_l$ scalar multiplications for the point-wise multiplication with $V_l$. Thus the structured approximation is only marginally more expensive. Further, it uses only $K_l + 2\times (K_{l-1}+1)$ weight variance parameters per layer, instead of $K_l \times (K_{l-1}+1)$ parameters used by the factorized approximation. Not having to compute gradients and update these additional parameters further mitigates the performance difference.
%
%
\subsection{Pruning Rule}
\label{sec:pr}
The Horseshoe and its regularized variant provide strong shrinkage towards zero for small $\wnode$. However, the shrunk weights, although tiny, are never actually zero. A user-defined thresholding rule is required to prune away the shrunk weights. One could first summarize the inferred posterior distributions using a point estimate and then use the summary to define a thresholding rule \citep{louizos2017bayesian}. We propose an alternate thresholding rule that obviates the need for a point summary. We prune away a unit, if $p(\taunode\taulayer < \delta) > p_0$, where $\delta$ and $p_0$ are user defined parameters, with $\taunode \sim q(\taunode\mid\phi_{\taunode})$ and $\taulayer \sim q(\taulayer\mid\phi_{\taulayer})$. Since, both $\taunode$ and $\taulayer$ are constrained to the log-Normal variational family, their product follows another log-Normal distribution, and implementing the thresholding rule simply amounts to computing the cumulative distribution function of the log-Normal distribution. To see why this rule is sensible, recall that for units which experience strong shrinkage the regularized Horseshoe tends to the Horseshoe. Under the Horseshoe prior, $\taunode\taulayer$ governs the (non-negative) scale of the weight node vector $\wnode$. Therefore, under our thresholding rule, we prune away nodes whose posterior scales, place probability greater than $p_0$ below a sufficiently small threshold $\delta$. In our experiments, we set $p_0 = 0.9$ and $\delta$ to either $10^{-3}$ or $10^{-5}$. 

\section{Related Work}
\vskip -0.03in
Bayesian neural networks have a long history. Early work can be traced back
to~\cite{WBuntine91, DMackay92, RNeal93}. These early approaches do not scale well to modern architectures or the large datasets required to learn them. Recent advances in stochastic MCMC methods~\cite{Cli2016, MWelling2011} and stochastic variational methods~\cite{CBlundell15, DRezende14}, black-box variational and alpha-divergence
minimization~\cite{MHLobato16, RRanganath14}, and probabilistic
backpropagation~\cite{MHLobato15} have reinvigorated
interest in BNNs by allowing scalable inference. 
\vskip -0.03in
Work on learning structure in BNNs has received less attention.    \cite{CBlundell15} introduce a mixture-of-Gaussians prior on the weights, with one mixture tightly concentrated around zero, thus approximating a spike and slab prior over weights. Others~\cite{DKingma15, YGal16Dropout} have noticed connections between Dropout~\cite{NSrivastava14} and approximate variational inference. In particular, \cite{DMolchanov17} show that the interpretation of Gaussian dropout as performing variational inference in a network with log uniform priors over weights leads to sparsity in weights. The goal of turning off edges is very different than the approach considered here, which performs model selection over the appropriate number of nodes. More closely related to us, are the recent works of~\cite{SGhosh17} and~\cite{louizos2017bayesian}. The authors consider group Horseshoe priors for unit pruning. We improve upon these works by using regularized Horseshoe priors that improve generalization, structured variational approximations that provide more accurate inferences, and by proposing a new thresholding rule to prune away units with small scales. Yet others~\cite{KNeklyudov2017structured} have proposed pruning units via truncated log-normal priors over unit scales. However, they do not place priors over network weights and are unable to infer posterior weight uncertainty. In related but orthogonal research~\cite{RAdams10, ZSong2017} focused on the problem of structure learning in deep belief networks.
 \vskip -0.03in
There is also a body of work on learning structure in non-Bayesian neural networks. Early work~\cite{lecun1990optimal, hassibi1993optimal}
pruned networks by analyzing second-order derivatives of the
objectives. More recently, \cite{wen2016learning} describe
applications of structured sparsity not only for optimizing filters
and layers but also computation time.  Closer to our work in spirit,
\cite{ochiai2016automatic, scardapane2017group, JAlvarez2016learning} and
\cite{murray2015auto} who use group sparsity to prune groups of
weights---e.g. weights incident to a node. However, these approaches
don't model weight uncertainty and provide uniform shrinkage
to all weights.  

\section{Experiments}
\label{sec:experiments}
\begin{figure*}[!ht]
\begin{center}
 \centerline{\includegraphics[width=0.8\linewidth]{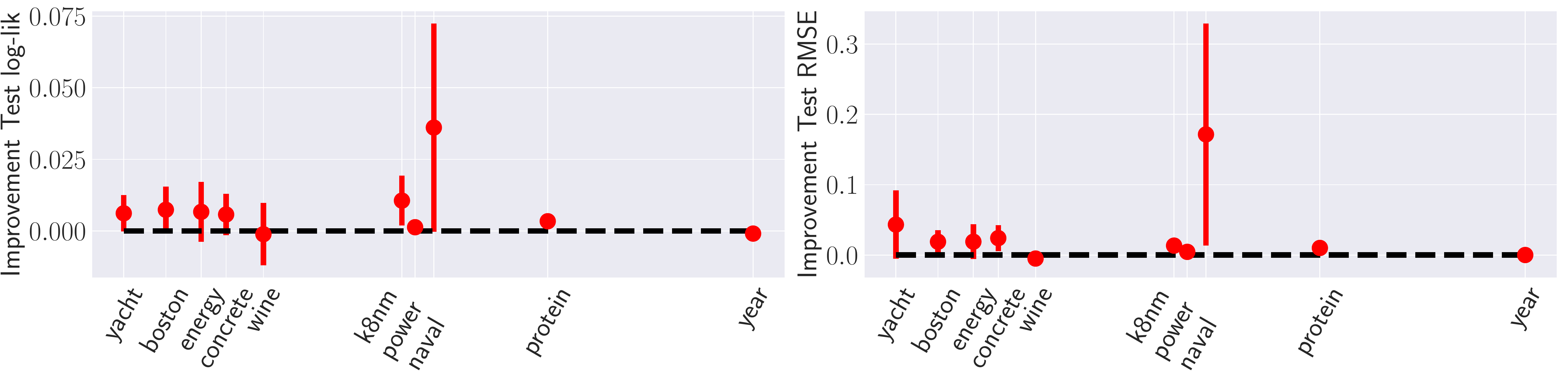}}
 \vskip -0.06in
 \centerline{\includegraphics[width=0.99\linewidth]{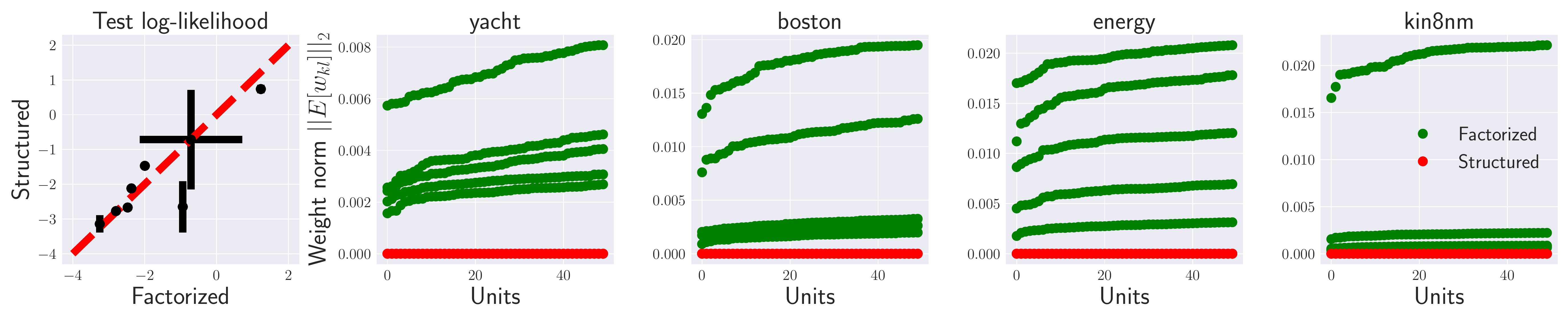}}
  \vskip -0.05in
 \centerline{\includegraphics[width=0.75\linewidth]{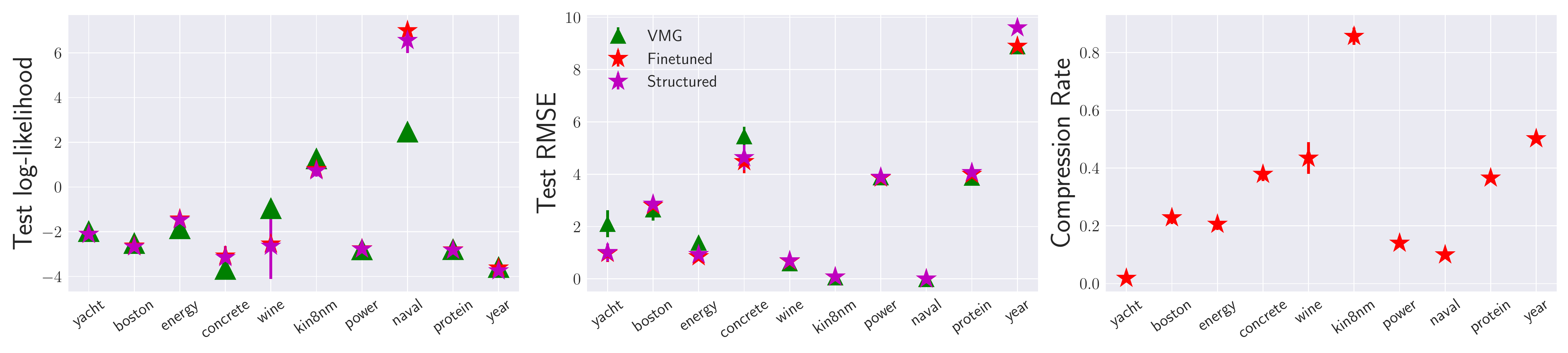}\includegraphics[width=0.24\linewidth]{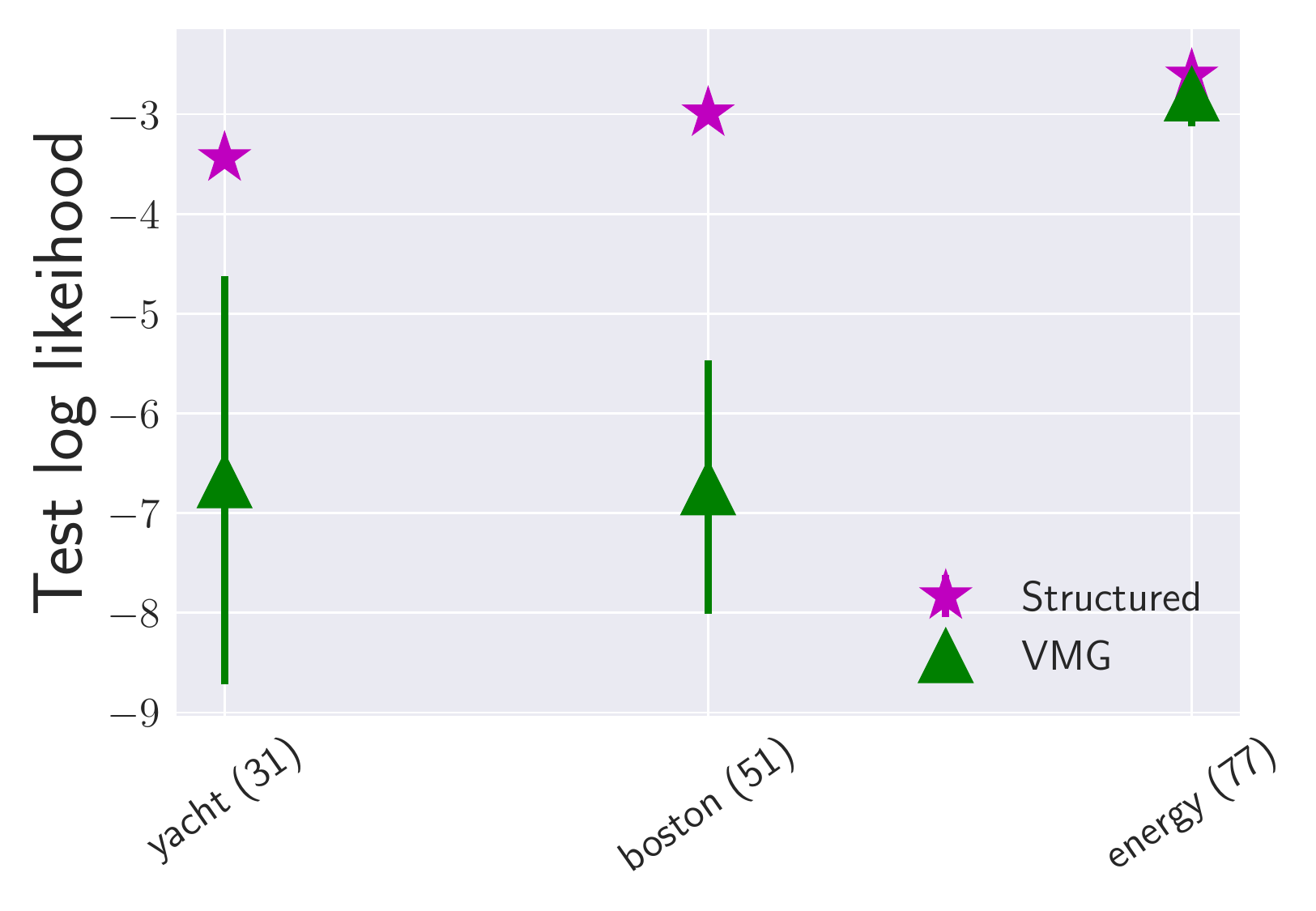}}
 \vskip -0.2in
\caption{\small{\emph{Top}: Regularized Horseshoe results in consistent improvements over the vanilla horseshoe prior. The datasets are sorted according to the number of data instances and plotted on the log scale with `yacht' being the smallest and `year' being the largest. Relative improvement is defined as $(x - y)/\textit{max}(|x|, |y|)$. \emph{Middle}: Structured variational approximations result in similar predictive performance but consistently recover solutions that exhibit stronger shrinkage. The left most figure plots the predictive log likelihoods achieved by the two approximations, each point corresponds to a UCI dataset. We also plot the fifty units with the smallest $||E[\wnode]||_2$, on a number of datasets. Each point in the plot displays the inferred $||E[\wnode]||_2$ for a unit in the network. We plot recovered expected weight norms from all five random trials for both the factorized and structured approximation. The structured approximation (in red) consistently provides stronger shrinkage. The factorized approximation both produces weaker shrinkage and the degree of shrinkage exhibits higher variance with random trials. \emph{Bottom}: The structured approximation is competitive with VMG while using much smaller networks.  Fine tuning occasionally leads to small improvements. Compression rates are defined as the fraction of un-pruned units. The rightmost plot compares VMG and reg-HS BNN in small data regimes on the three smallest UCI datasets. In parenthesis we indicate the number of training instances. The shrinkage afforded by reg-HS leads to improved performance over VMG which employs priors that lack shrinkage towards zero.}}
\vskip -0.3in
\label{fig:reghs}
\end{center}
\end{figure*}

\vskip -0.05in
In this section, we present experiments that evaluate various aspects of the proposed regularized Horseshoe Bayesian neural network (reg-HS) and the structured variational approximation. In all experiments, we use a learning rate of $0.005$, the global horseshoe scale $\blayer = 10^{-5}$, a batch size of $128$, $c_a=2$, and $c_b=6$. For the structured approximation, we also found that  constraining $\Psi$, $V$, and $h$ to unit-norms resulted in better predictive performance. 
Additional experimental details are in the supplement. 
\paragraph{Regularized Horseshoe Priors provide consistent benefits, especially on smaller data sets.}
We begin by comparing reg-HS against BNNs using the standard Horseshoe (HS) prior on a collection of diverse datasets from the UCI repository. We follow the protocol of~\cite{MHLobato15} to compare the two models. To provide a controlled comparison, and to tease apart the effects of model versus inference enhancements we employ factorized variational approximations for either model.  In figure~\ref{fig:reghs}, the UCI datasets are sorted from left to right, with the smallest on the left. We find that the regularized Horseshoe leads to consistent improvements in predictive performance. As expected, the gains are more prominent for the smaller datasets for which the regularization afforded by the regularized Horseshoe is crucial for avoiding over-fitting. In the remainder, all reported experimental results use the reg-HS prior.
\paragraph{Structured variational approximations provide greater shrinkage.} 
Next, we evaluate the effect of utilizing structured variational approximations. In preliminary experiments, we found that of the approximations described in Section~\ref{sec:v_families}, the structured approximation outperformed the semi-structured variant while the factorized approximation provided better predictive performance than the tied approximation. In this section we only report results comparing models employing these two variational families.

\begin{figure}[ht]
\begin{center}
 \centerline{\includegraphics[width=0.9\columnwidth]{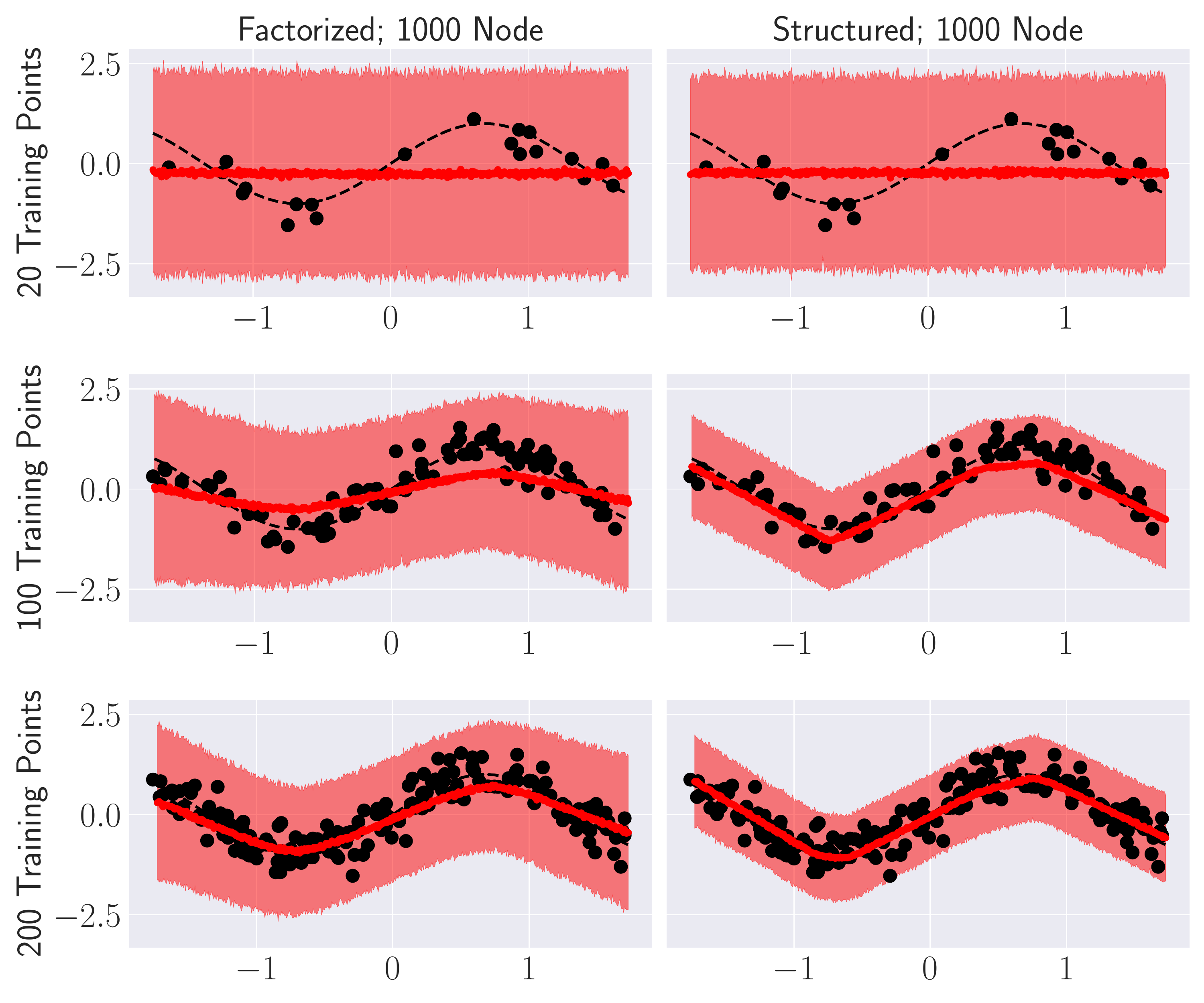}}
 \vskip -0.2in
\caption{\small{Regularized Horseshoe BNNs prune away excess capacity and are more resistant to underfitting. Variational approximations aware of model structure improve fits.}}
\label{fig:hs_bnn}
\end{center}
\vskip -0.4in
\end{figure}

\emph{Toy Data}
First, we explore the effects of structured and factorized variational approximations on predictive uncertainties. Following~\cite{SGhosh17} we consider a noisy regression problem: $y = sin(x) + \epsilon$, $\epsilon \sim \mathcal{N}(0, 0.1)$, and explore the relationship between predictive uncertainty and model capacity. We compare a single layer $1000$ unit BNN using a standard normal prior against BNNs with the regularized horseshoe prior utilizing factorized and structured variational approximations.  Figures~\ref{fig:uncertainty} and~\ref{fig:hs_bnn} show that while a BNN severely over-estimates the predictive uncertainty, models using the reg-HS priors  by pruning away excess capacity, significantly improve the estimated uncertainty. Furthermore, we observe that the structured approximation best alleviates the under-fitting issues.  
%
%

\emph{Controlled comparisons on UCI benchmarks}
We return to the UCI benchmark to carefully vet the different variational approximations. We deviate from prior work, by using networks with significantly more capacity than previously considered for this benchmark. In particular, we use single layer networks with an order of magnitude more hidden units ($500$) than considered in previous work ($50$). This additional capacity is more than that needed to explain the UCI benchmark datasets well. With this experimental setup, we are able to evaluate how well the proposed methods perform at pruning away extra modeling capacity. For all but the `year` dataset, we report results from five trials each trained on a random $90/10$ split of the data. For the large year dataset, we ran a single trial (details in the supplement).  Figure~\ref{fig:reghs} shows consistently stronger shrinkage.  

\emph{Comparison against Factorized approximations.}
The factorized and structured variational approximations have similar predictive performance. However, the structured approximation consistently recovers solutions that exhibit much stronger shrinkage towards zero. Figure~\ref{fig:reghs} demonstrates this effect on several UCI datasets, with more in the supplement. We have plotted $50$ units with the smallest $||\wnode||_2$ weight norms recovered by the factorized and structured approximations, from five random trials. Both approximations provide shrinkage towards zero, but the structured approximation has significantly stronger shrinkage. Further, the degree of shrinkage from the factorized approximation varies significantly between random initializations. In contrast, the structured approximation \emph{consistently} provides strong shrinkage. We compare the shrinkages using $||\E{\wnode}||_2$ instead of applying the pruning rule from section~\ref{sec:pr} and comparing the resulting compression rates. This is because although the scales $\taunode\taulayer$ inferred by the factorized approximation provide a clear separation between signal and noise, they do not exhibit shrinkage toward zero. However, $\wnode = \taunode\taulayer\betanode$ does exhibit shrinkage and provides a fair comparison.

\emph{Comparison against competing methods.}
We compare the reg-HS model with structured variational approximation against the variational matrix Gaussian (VMG) approach of~\cite{CLouizos16},  which has previously been shown to outperform other variational approaches to learning BNNs. We used the pruning rule with $\delta = 10^{-3}$ for all but the `year` dataset, for which we set $\delta=10^{-5}$. Figure~\ref{fig:reghs} demonstrates that structured reg-HS is competitive with VMG in terms of predictive performance. We either perform similarly or better than VMG on the majority of the datasets. More interestingly, structured reg-HS achieves competitive performance while pruning away excess capacity and achieving significant compression. We also fine-tuned the pruned model by updating the weight means while holding others fixed. However, this didn't significantly affect predictive performance. Finally, we evaluate how reg-HS compares against VMG in the low data regime. For the three smallest UCI datasets we use ten percent of the data for training. In such limited data regimes (Figure~\ref{fig:reghs}) the shrinkage afforded by reg-HS leads to  clear improvements in predictive performance over VMG.

\paragraph{HS-BNNs improve reinforcement learning performance.}
So far, we have focused on using BNNs simply for prediction.  One application area in which having good predictive uncertainty estimates is crucial is in model-based reinforcement learning scenarios (e.g. \citep{depeweg2016learning, gal2016improving,killian2017robust}): here, it is essential not only to have an estimate of what state an agent may be in after taking a particular action, but also an accurate sense of all the states the agent may end up in. In the following, we apply our regularized HS-BNN with structured approximations to two domains: the 2D map of \citet{killian2017robust} and acrobot \citet{sutton1998reinforcement}. For each domain, we focused on one instance dynamic setting. In each domain, we collected training samples by training a DDQN~\citep{van2016deep} online (updated every episode). The DDQN was trained with an epsilon-greedy policy that started at one and decayed to $0.15$ with decay rate $0.99$, for 500 episodes. This procedure ensured that we had a wide variety of samples that were still biased in coverage toward the optimal policy. To simulate resource constrained scenarios,  we limited ourselves to 10$\%$ of DDQN training batches ($346$ samples for the 2D map and $822$ training samples for acrobot). We considered two architectures, a single hidden layer network with $500$ units, and a two layer network with $100$ units per layer as the transition function for each domain.  Then we simulated from each BNN to learn a DDQN policy (two layers of width $256$, $512$; learning rate $5e-4$) and tested this policy on the original simulator.  

As in our prediction results, training a moderately-sized BNN with so few data results in severe underfitting, which in turn, adversely affects the quality of the policy that is learned.  We see in table~\ref{tab:rl} that the better fitting of the structured reg-HS-BNN results in higher task performance, across domains and model architectures.
\vskip -0.2in
\begin{table}[!h] 
  \caption{Model-based reinforcement learning.  The under-fitting of the standard BNN results in lower task performance, whereas the HS-BNN is more robust to this underfitting.}
\begin{center}
\resizebox{\columnwidth}{!}{%
\begin{tabular}{|l|l|l|} \hline 
& \multicolumn{2}{|c|}{2D Map}\\ \hline 
   & Test RMSE  & Avg. Reward\\ \hline
  BNN x-500-y &  0.187 & 975.386 \\\hline
  BNN x-100-100-y &  0.089 & 966.716 \\\hline
  \textbf{Structured x-500-y} & \textbf{0.058} & \textbf{995.416} \\\hline
  Structured x-100-100-y & 0.061 & 992.893 \\\hline
& \multicolumn{2}{|c|}{Acrobot}\\ \hline 
  BNN x-500-y & 0.924 & -156.573 \\\hline
  BNN x-100-100-y & 0.710 & -23.419 \\\hline
  Structured  x-500-y & \textbf{0.558} & -108.443 \\\hline
  \textbf{Structured  x-100-100-y} & 0.656 & \textbf{-17.530} \\\hline
  \end{tabular}%
  } 
\end{center}
  \label{tab:rl}
\end{table} 
\vspace{-0.2in}

\section{Discussion and Conclusion}
We demonstrated that the regularized
horseshoe prior, combined with a structured variational 
distribution, is a computationally efficient tool for model selection
in Bayesian neural networks. By retaining crucial posterior dependencies, the structured approximation provided, to our knowledge, state of the
art shrinkage for BNNs while being competitive in predictive performance to  existing approaches.
We found, model re-parameterizations --- decomposition of the Half-Cauchy priors into inverse gamma distributions and non-centered representations essential for avoiding poor local optima.
There remain several interesting follow-on directions, including, modeling enhancements that use layer, node, or even weight specific weight decay $c$, or layer specific global shrinkage parameter $b_g$ to provide different levels of shrinkage to different parts of the BNN.

\bibliography{egbib.bib}
\bibliographystyle{icml2018}
\newpage
\appendix
\onecolumn
\section{Conditional variational pre-activations}
Recall from Section 4.2, that the variational pre-activation distribution is given by $q(b \mid a, \nu_l, \phi_{\betamat}) = \normal(b \mid \mu_b, \Sigma_b)
 = \normal(b \mid \cbetamat^T a, (a^T\cU a)V)$, where $U = \Psi + hh'$, and $V$ is diagonal.
 To equation requires $\cbetamat$ and $\cU$. The expressions for these follow directly from the properties of partitioned Gaussians. 

For a particular layer $l$, we drop the explicit dependency on $l$ from the notation. Recall that $ B=
\left[
   \begin{array}{c}
     \beta \\
	\nu^T
    \end{array}
\right ]$, and let $B \in \real{m \times n}$, $\beta \in \real{m-1 \times n}$, and $\nu \in \real{n \times 1}$ $q(B\mid\phi_{B}) = \mvn(B\mid M, U, V)$. From properties of the Matrix normal distribution, we know that a column-wise vectorization of $B$, $\vecB \sim \normal(\vecM, V\otimes U)$. From this and Gaussian marginalization properties it follows that the $j^{\text{th}}$ column $t_j = [\beta_j; \nu_j]$ of $B$ is distributed as $t_j \sim \normal(m_j, V_{jj}U)$, where $m_j$ is the appropriate column of $M$. Conditioning on $\nu_j$ then yields, $q(\beta_j \mid \nu_j) = \normal(\beta_j \mid \mu_{\cbetaj}, \Sigma_{\cbetaj})$, where 
\begin{equation}
\begin{split}
	&\Sigma_{\cbetaj} 
	= V_{jj} (\Psi_{\beta} + \frac{\Psi_{\nu}}{\Psi_{\nu} + h_{\nu}^2}h_{{\beta}}h_{{\beta}}^T) \\
	&\mu_{\cbetaj} = \mu_{{\beta_j}} + \frac{h_{\nu}(\nu_j - \mu_{{\nu_j}})}{\Psi_{\nu} + h_{\nu}^2} h_{\beta}
\end{split}
\end{equation}
Rearranging, we can see that, $M_{\beta\mid\nu}$ is made up of the columns $\mu_{\cbetaj}$ and $U_{\beta\mid\nu} = \Psi_{\beta} + \frac{\Psi_{\nu}}{\Psi_{\nu} + h_{\nu}^2}h_{\beta}h_{\beta}^T$.

\section {Algorithmic details}
The ELBO corresponding to the non-centered regularized HS model is,
\begin{equation}
\begin{split}
	\elbo(\phi) &= \E{\text{ln }\invgamma(c \mid c_a, c_b)} + \E{\text{ln }\invgamma(\kappa \mid 1/2, 1/\lambdakappa)} + \E{\text{ln }\invgamma(\lambdakappa \mid 1/2, 1/\bkappa^2)}  \\
	&+\sum_n \E{\text{ln } p(y_n \mid \beta, \mathcal{T}, \kappa, x_n)} \\
	&+\sum_{l=1}^{L-1}\sum_{k=1}^{K_L} 
	 \E{\text{ln }\invgamma(\lambdanode \mid 1/2, 1/\bnode^2)}\\
	  &+\sum_{l=1}^{L-1} \E{\text{ln }\invgamma(\taulayer\mid 1/2, 1/\lambdalayer)} + \E{\text{ln }\invgamma(\lambdalayer\mid 1/2, 1/\blayer^2)}\\
	   &+\sum_{l=1}^{L-1}\Ewrt{q(B_l)}{\text{ln }\normal(\beta_{l}\mid 0, \eye) + \text{ln }\invgamma(\tau_l \mid 1/2, 1/\lambda_l)  } + \sum_{k=1}^{K_L} \E{\text{ln }\normal(\beta_{kL}\mid 0, \eye )} + \ent{q(\theta\mid \phi)}.
\end{split}
\end{equation}
We rely on a Monte-Carlo estimates to evaluate the expectation involving the likelihood $\E{\text{ln } p(y_n \mid \beta, \mathcal{T}, \kappa, x_n)}$. 
\paragraph{Efficient computation of the Matrix Normal Entropy}
The entropy of $q(B) = \mvn(B\mid M, U, V)$ is given by $\frac{mn}{2}\text{ln }(2\pi e) + \frac{1}{2}\text{ln }|V\otimes U|$. We can exploit the structure of $U$ and $V$ to compute this efficiently. We note that $\text{ln } |V \otimes U| = m\text{ln }|V| + n\text{ln }|U|$. Since $V$ is diagonal $\text{ln }|V| = \sum_j \text{ln } V_{jj}$. Using the matrix determinant lemma we can efficiently compute $|U| = (1 + h'\Psi^{-1}h)|\Psi|$. Owing to the diagonal structure of $\Psi$, computing it's determinant and inverse is particularly efficient. 
\paragraph{Fixed point updates}
The auxiliary variables $\lambdakappa$, $\lambdalayer$ and $\lambdalayer$ all follow inverse Gamma distributions. Here we derive for $\lambdanode$, the others follow analogously.
Consider,
\begin{equation}
\begin{split}
	\text{ln } q(\lambdanode) &\propto \Ewrt{-q_{\lambdanode}}{\text{ln }\invgamma(\taunode \mid 1/2, 1/\lambdanode)} + \Ewrt{-q_{\lambdanode}}{\text{ln }\invgamma(\lambdanode \mid 1/2, 1/\bnode^2)}, \\
	&\propto (-1/2 -1/2 -1)\text{ln } \lambdanode - (\E{1/\taunode} + 1/\bnode^2)(1/\lambdanode),
	\end{split}
\end{equation}
from which we see that,
\begin{equation}
\begin{split}
q(\lambdanode) = \invgamma(\lambdanode \mid c, d), \\
c = 1, d = \E{\frac{1}{\taunode}} + \frac{1}{\bnode^2}.	
\end{split}
\label{eqn:fxp}
\end{equation}
Since, $q(\taunode) = \text{ln }\normal(\mu_{\taunode}, \sigma^2_{\taunode})$, it follows that $\E{\frac{1}{\taunode}} = \text{exp}\{-\mu_{\taunode} + 0.5*\sigma^2_{\taunode}\}$. We can thus calculate the necessary fixed point updates for $\lambdanode$ conditioned on $\mu_{\taunode}$ and $\sigma^2_{\taunode}$. Our algorithm uses these fixed point updates given estimates of $\mu_{\taunode}$ and $\sigma^2_{\taunode}$ after each Adam step.
\section{Algorithm}
Algorithm~\ref{alg:overall} provides pseudocode summarizing the overall algorithm for training regularized HSBNN (with strictured variational approximations).
\begin{algorithm}[!h]
  \caption{Regularized HS-BNN Training}
  \label{alg:overall}
  \begin{algorithmic}[1]
    \STATE \textbf{Input} Model $p(\data, \theta)$, variational approximation  $q(\theta\mid\phi)$, number of iterations T.
    \STATE \textbf{Output}: Variational parameters $\phi$  
    \STATE Initialize variational parameters $\phi$.
    \FOR{\texttt{T iterations}}
    \STATE Update $\phi_c$, $\phi_\kappa$, $\phi_\gamma$, $\{\phi_{B_l}\}_l$, $\{\phi_{\taulayer}\}_l \leftarrow \text{ADAM}(\elbo(\phi))$.
    \FOR{\texttt{all hidden layers} $l$}
    \STATE Conditioned on $\phi_{B_l}$, $\phi_{\taulayer}$ update $\phi_{\lambdalayer}$, $\phi_{\lambdanode}$ using fixed point updates (Equation~\ref{eqn:fxp}).
    \ENDFOR
    \STATE Conditioned on $\phi_\kappa$ update $\phi_{\lambdakappa}$ via the corresponding fixed point update.
    \ENDFOR
\end{algorithmic}
\end{algorithm}
\section{Experimental details}
For regression problems we use Gaussian likelihoods with an unknown precision $\gamma$, $p(y_n \mid f(\W, x_n), \gamma) = \normal(y_n \mid f(\W, x_n), \gamma^{-1})$. We place a vague prior on the precision,$\gamma \sim \text{Gamma}(6, 6)$ and approximate the posterior over $\gamma$ using another variational distribution $q(\gamma \mid \phi_\gamma)$. The corresponding variational parameters are learned via a gradient update during learning. 

\paragraph{Regression Experiments}
 For comparing the reg-HS and HS models we followed the protocol of (Hernandez-Lobato \& Adams, 2015) and trained a single hidden layer network with $50$ rectified linear units for all but the larger ``Protein'' and ``Year'' datasets for which we train a $100$ unit network. For the smaller datasets we train on a randomly subsampled $90\%$ subset and evaluate on the remainder and repeat this process $20$ times. For ``Protein'' we perform 5 replications and for ``Year'' we evaluate on a single split. For, VMG we used $10$ pseudo-inputs, a learning rate of $0.001$ and a batch size of 128.
 
 \paragraph{Reinforcement learning Experiments} We used a learning rate of $2e-4$. For the $2D$ map domain we trained for $1500$ epochs and for acrobot we trained for $2000$ epochs. 
\vskip 2.5in
\section{Additional Experimental results}
\begin{figure*}[ht]
\includegraphics[width=1\textwidth]{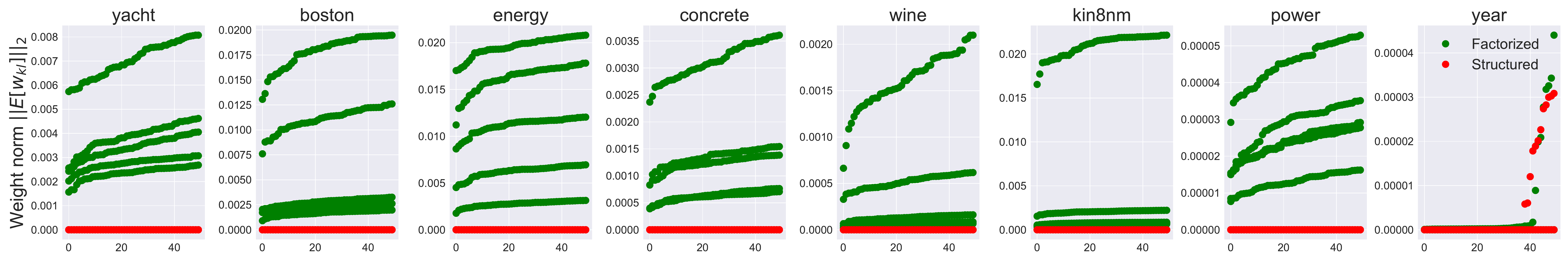}
\vskip -0.2in
\caption{Structured variational approximations consistently recover solutions that exhibit stronger shrinkage. We plot the 50 smallest $||\wnode||_2$ recovered by the two approximations on five random trials on a number of UCI datasets. }	
\label{fig:shrink}
\end{figure*} 

In Figure~\ref{fig:shrink} we provide further shrinkage results from the experiments described in the main text comparing regularized Horseshoe models utilizing factorized and structured approximations.

\begin{figure}[ht]
\centering
\includegraphics[width=0.95\textwidth]{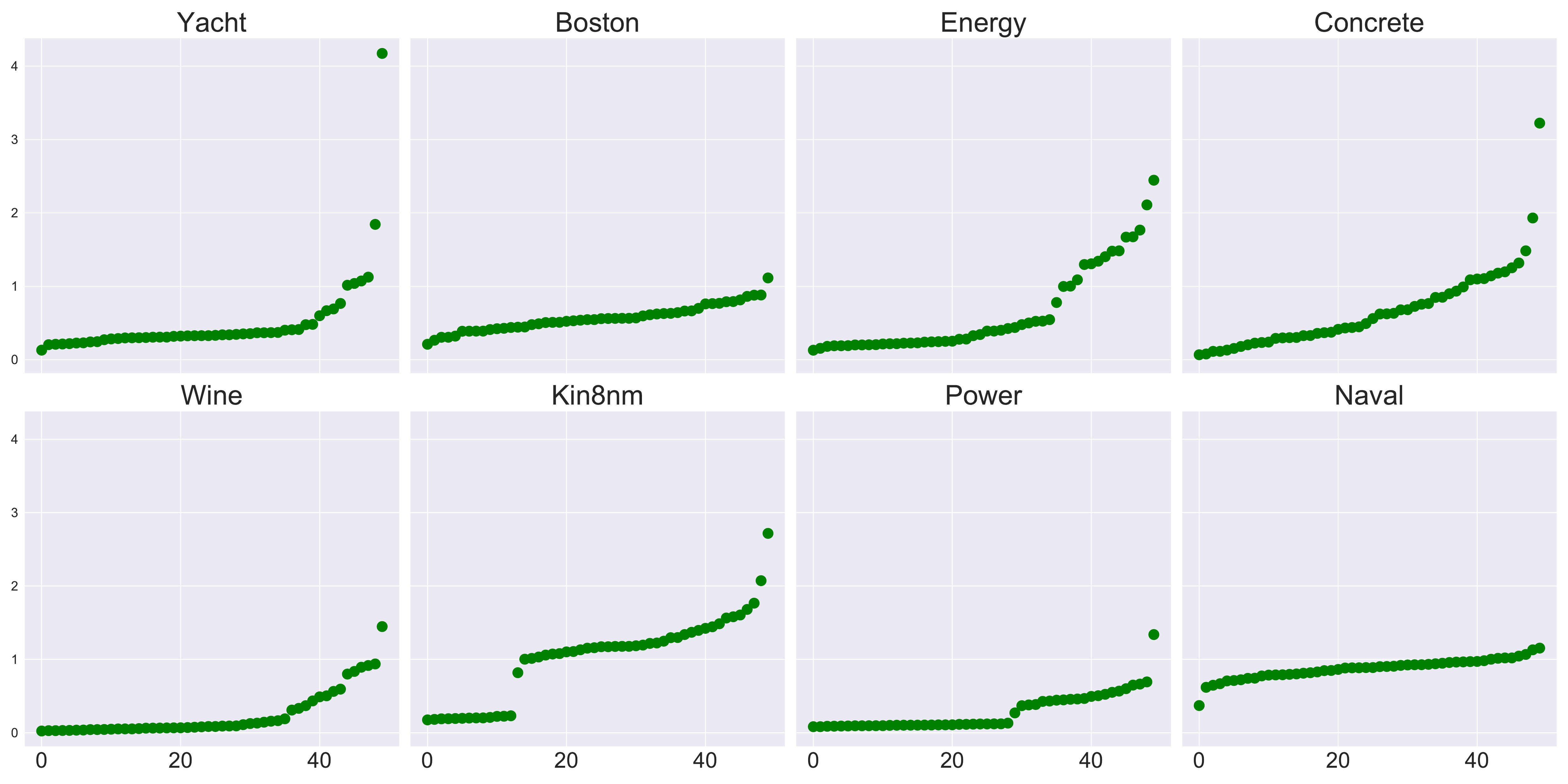}
\vskip -0.1in
\caption{ We plot $||\wnode||_2$ recovered by the HS-BNN using the fully factorized variational approximation on a number of UCI datasets. }	
\label{fig:sup2}
\end{figure}

\paragraph{Shrinkage provided by fully factorized Horseshoe BNNs on UCI benchmarks}
Figure~\ref{fig:sup2} illustrates the shrinkage afforded by 50 unit HS-BNNs using fully factorized approximations. Similar to factorized regularized Horseshoe BNNs limited compression is achieved. Figures On some datasets, we do not achieve much compression and all 50 units are used. A consequence of the fully factorized approximations providing weaker shrinkage as well as $50$ units not being large enough to model the complexity of the dataset.

\newpage
\section{Prior samples from networks with HS and regularized Horseshoe priors}
\begin{figure}[ht]
\centering
\includegraphics[width=1.0\textwidth]{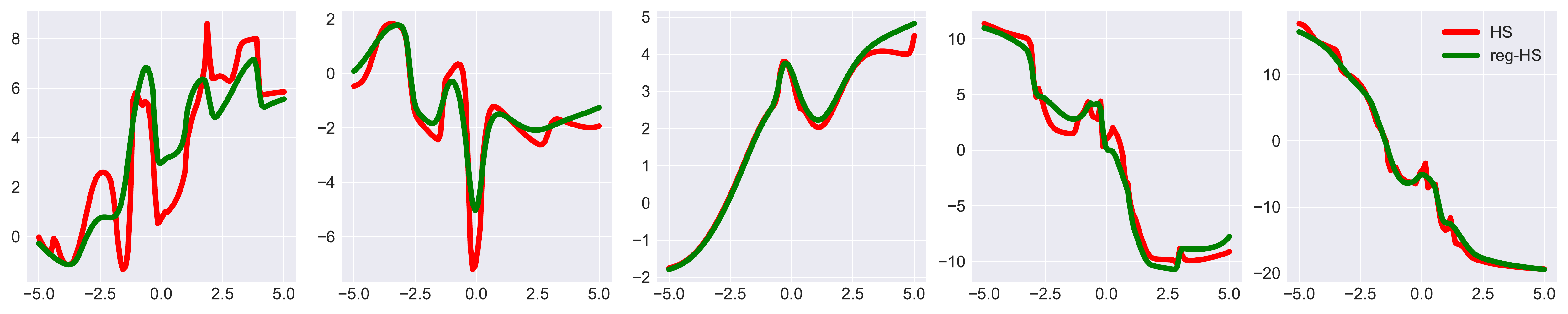}
\includegraphics[width=1.0\textwidth]{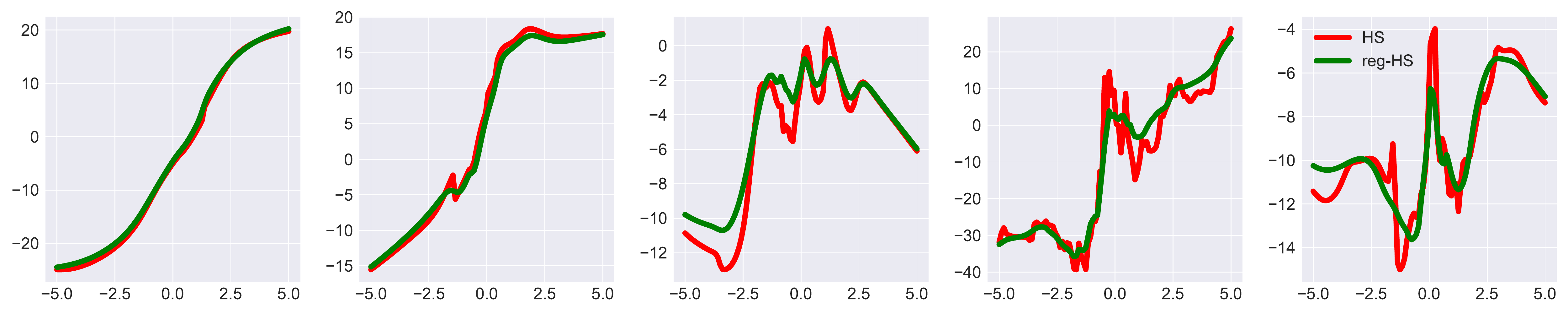}
\includegraphics[width=1.0\textwidth]{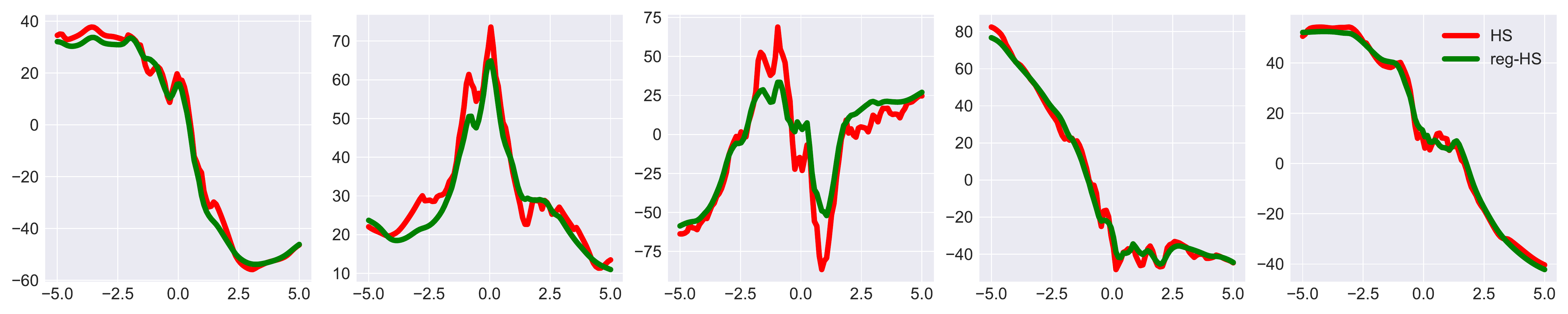}
\vskip -0.1in
\caption{ Each row displays five random samples drawn from a single layer network with \textsc{tanh} non-linearities. The top row contains samples from a $50$ unit network, the middle row contains samples from a $500$ unit network and the bottom row displays samples from a $5000$ unit network. Matched samples from the regularized HS and HS priors were generated by sharing $\betanode$ samples between the two. The hyper-parameters used were $b_0$ = $b_g = 1$, $c_a = 2$ and $c_b = 6$.}	
\label{fig:sup3}
\end{figure}

To provide further intuition into the behavior of networks with Horseshoe and regularized Horseshoe priors we provide functions drawn from networks endowed with these priors. Figure~\ref{fig:sup3} plots five random functions sampled from one layer networks with varying widths. Observe that the regularized horseshoe distribution leads to smoother functions, thus affording stronger regularization. As demonstrated in the main paper, this stronger regularization leads to improved predictive performance when the amount of training data is limited.

\end{document}